\documentclass{article}
\usepackage{ijcai24}

\usepackage{times}
\usepackage{soul}
\usepackage{url}
\usepackage[hidelinks]{hyperref}
\usepackage[utf8]{inputenc}
\usepackage[small]{caption}
\usepackage{graphicx}
\usepackage{amsmath}
\usepackage{amsthm}
\usepackage{amssymb}
\usepackage{booktabs}
\usepackage[ruled]{algorithm2e}
\SetKwComment{Comment}{// }{}
\usepackage[switch]{lineno}

\title{Detecting and Understanding Vulnerabilities in Language Models via Mechanistic Interpretability}

\author{
Jorge García-Carrasco
\and
Alejandro Maté\And
Juan Trujillo\\
\affiliations
Lucentia Research, Department of Software and Computing Systems, University of Alicante\\
\emails
jorge.g@ua.es,
\{amate, jtrujillo\}@dlsi.ua.es,
}

\begin{document}

\maketitle

\begin{abstract}

    Large Language Models (LLMs), characterized by being trained on broad amounts of data in a self-supervised manner, have shown impressive performance across a wide range of tasks. Indeed, their generative abilities have aroused interest on the application of LLMs across a wide range of contexts. However, neural networks in general, and LLMs in particular, are known to be vulnerable to adversarial attacks, where an imperceptible change to the input can mislead the output of the model. This is a serious concern that impedes the use of LLMs on high-stakes applications, such as healthcare, where a wrong prediction can imply serious consequences. Even though there are many efforts on making LLMs more robust to adversarial attacks, there are almost no works that study \emph{how} and \emph{where} these vulnerabilities that make LLMs prone to adversarial attacks happen. Motivated by these facts, we explore how to localize and understand vulnerabilities, and propose a method, based on Mechanistic Interpretability (MI) techniques, to guide this process. Specifically, this method enables us to detect vulnerabilities related to a concrete task by (i) obtaining the subset of the model that is responsible for that task, (ii) generating adversarial samples for that task, and (iii) using MI techniques together with the previous samples to discover and understand the possible vulnerabilities. We showcase our method on a pretrained GPT-2 Small model carrying out the task of predicting 3-letter acronyms to demonstrate its effectiveness on locating and understanding concrete vulnerabilities of the model. 
 
\end{abstract}

\section{Introduction}

Scaling up the size of Large Language Models (LLMs) is giving impressive performance on a wide range of tasks \cite{brown2020language}. In fact, there is strong empirical evidence supporting that the performance of language models, based on the Transformer architecture \cite{vaswani2017attention}, is directly related to the number of parameters, dataset size and training computation \cite{kaplan2020scaling}. Hence, the general abilities of LLMs, including reasoning, are expected to keep on improving. Therefore, there is an increasing interest on applying LLMs for high-stakes applications, such as healthcare \cite{cascella2023evaluating}. 

However, it has been shown that neural networks in general, and LLMs in particular, are vulnerable to adversarial attacks \cite{shayegani2023survey}. An adversarial attack consists on slightly perturbing the input so that it is misclassified by the model \cite{huang2020survey}. For example, changing a single word can cause models such as GPT-2 or BERT to output a completely different classification \cite{guo-etal-2021-gradient}. This can be a serious concern regarding the applicability of LLMs on tasks where a wrong prediction or behavior can have serious consequences. 

There are many works that focus on increasing the robustness to face such adversarial attacks. Most of these works are based on the so-called adversarial training scheme \cite{DBLP:journals/corr/GoodfellowSS14,madry2018towards,wang2019improving,liu2020adversarial}, which essentially consists on including adversarial examples into the training scheme so that the model is ``aware" of the possible vulnerabilities that it might have. 

Nevertheless, and to the best of our knowledge, the current state of the art has not devoted enough attention in order to try to localize and understand the underlying mechanism behind a certain vulnerability. In other words, the question “Is it possible to locate the exact components of the model that are affected by a certain vulnerability?” has not yet been studied in detail. In the affirmative case, Can the vulnerability be understood in terms of the behavior of such components? Understanding the mechanism under a vulnerability would enable us to gauge our trustworthiness on a model, and even being able to mitigate the effects of such vulnerability.


When it comes to these questions, there is evidence that concrete behaviors of LLMs can be explained in terms of a small subset of components. In other words, specific abilities of an LLM (e.g. the ability to detect capital letters) can be \emph{located} on a small subset of components. Detecting and understanding these components is one of the main research topics related to Mechanistic Interpretability (MI), which is a recent field that tries to interpret the behavior of neural networks in terms of human-understandable concepts \cite{elhage2021mathematical,elhage2022toy,olsson2022context}. The main approach is to employ a set of causal interventions to isolate a \emph{circuit} \cite{olah2020zoom} (i.e. a subset of the model) that is responsible for a concrete task. For example, \cite{wang2022interpretability} discover the circuit responsible for the Indirect Object Identification task (IOI) in GPT-2
Small. Similarly, \cite{hanna2023does} use MI techniques to explain how GPT-2 Small performs the greater-than operation on a single task and test if the discovered circuit generalizes to other contexts, whereas \cite{pmlr-v238-garcia-carrasco24a} studied the acronym prediction task. Likewise, \cite{docstring} discovered how a smaller 4-layer transformer model predicted argument names on a docstring. 

In summary, the aforementioned works show that it is possible to locate and understand the circuit that explains a specific behavior of a language model. While understanding such circuit can help in detecting the corresponding possible vulnerabilities, the current approach is to manually inspect the circuit for potential weaknesses, which can be time-consuming and requires a large degree of experience. Therefore, even though current MI techniques can be used to locate and understand circuits, there are no works that explore how to systematically identify and understand vulnerabilities in such circuits.

Motivated by these facts, our work proposes an approach to identify which components of the model are vulnerable on a specific task, and the mechanism behind such vulnerability. Specifically, given a specific task, we (i) identify and understand the circuit associated with that task via MI techniques, (ii) automatically generate adversarial samples related to the task of study and (iii) use the adversarial samples to locate the exact components of the circuit that are vulnerable to such attack and understand why does it happen. 

Summarizing, the main contributions of our work are:

\begin{itemize}
    \item The proposal of a new method to detect and understand vulnerabilities in language models. Given a concrete task or behavior, our approach enables us to systematically detect and understand the possible vulnerabilities associated to the underlying circuit.
    \item A showcase of our proposal on a case study to locate and understand vulnerabilities on the task of 3-letter acronym prediction using GPT-2 Small.
\end{itemize} 

To the best of our knowledge, this is the first work that analyzes vulnerabilities of models from the MI perspective. We believe that zooming into the internals of the models can give valuable insights that can help us detect, understand and, in the future, palliate or even solve these vulnerabilities without requiring extra adversarial training or risking inadvertently causing collateral effects. 

The rest of the paper is structured as follows: Section \ref{sec:background} presents the required background. Section \ref{sec:approach} describes our approach to locate and understand vulnerabilities. Section \ref{sec:cases} showcases the approach by applying it to detecting vulnerabilities on GPT-2 Small. Section \ref{sec:discussion} discusses about the approach and its application to the case of study. Finally, the conclusions are presented in \ref{sec:conclusion}.

\section{Background}
\label{sec:background}

In this section, we briefly present the necessary background and techniques that will be used to detect and understand vulnerabilities in language models.

\subsection{Model and Notation}

The GPT-2 Small model \cite{radford2019language}, is a decoder-only transformer architecture with 117 million parameters. It consists of 12 transformer blocks, each containing 12 attention heads, followed by a Multi-Layer Perceptron (MLP). Layer Normalization \cite{ba2016layer} that is applied to each component. The model takes an input sequence of $N$ consecutive tokens, which are embedded into $x_0 \in \mathbb{R}^{N \times d}$ using a learned embedding matrix $W_E \in \mathbb{R}^{V \times d}$, where $V$ represents the vocabulary size. Additionally, positional embeddings are added to $x_0$.

As shown by \cite{elhage2021mathematical}, interpreting a transformer architecture is easier when thinking of it as a \emph{residual stream}, whose initial value is $x_0$, where all the components sequentially read from and write to, modifying the initial vector. Finally, the final residual vector is unembedded using a matrix $W_U$, which is tied to the embedding matrix (i.e., $W_U = W_E^T$) in the case of GPT-2. This unembedding process yields a vector $y \in \mathbb{R}^{N \times V}$, where $y_{ij}$ represents the logits for the $j$th token in the vocabulary, following the prediction of the $i$th token in the sequence.

\subsection{Logit Attribution}

As previously-seen, the logits are obtained by linearly mapping the final residual stream vector by using the unembedding matrix. As layer normalization can also be approximated as a linear map, it implies that the logits can be decomposed as a sum of contributions of the different components. Essentially, this can be used to analyze which components contribute the most to the correct (or incorrect) prediction. More formally, if $h_{ij}$ is the output of the $i$th attention head of the $j$th layer, the logit attribution on the $k$th token can be expressed as:

\begin{equation}
    \text{logit\_attr}_k (h_{ij}) = W_U [k] \cdot h_{ij} 
\end{equation}

where $W_U [k]$ is the $k$th column of the unembedding matrix.

\subsection{Activation Patching}

Activation patching, first presented in \cite{meng2022locating}, consists on \emph{patching} (i.e. replacing) the activations of a given component with the activations obtained by running the model on a \emph{corrupted prompt}. If patching the activation of a given component causes a large drop of performance, it implies that such component is relevant to the task of study, hence enabling us to locate the circuit. 

When it comes to corrupting the prompt, many ways have been used across the literature. For example, it is common to perform \emph{zero ablation} \cite{mcgrath2023hydra}, which consists on simply setting the activations of a given component to zero, or \emph{noise ablation} \cite{meng2022locating}, which adds noise to the activations sampled from a Gaussian distribution. However, these methods modify the activations such that the model goes off-distribution, which can give misleading results. In our approach, we use \emph{resample ablation} \cite{causalscrubbing}, which simply replaces the activations with other activations from different prompts of the dataset. This process is more principled as the others, as it uses activations that are in-distribution.

\section{Our approach}
\label{sec:approach}

\begin{figure*}[!tb]
    \centering
    \includegraphics[width=\linewidth]{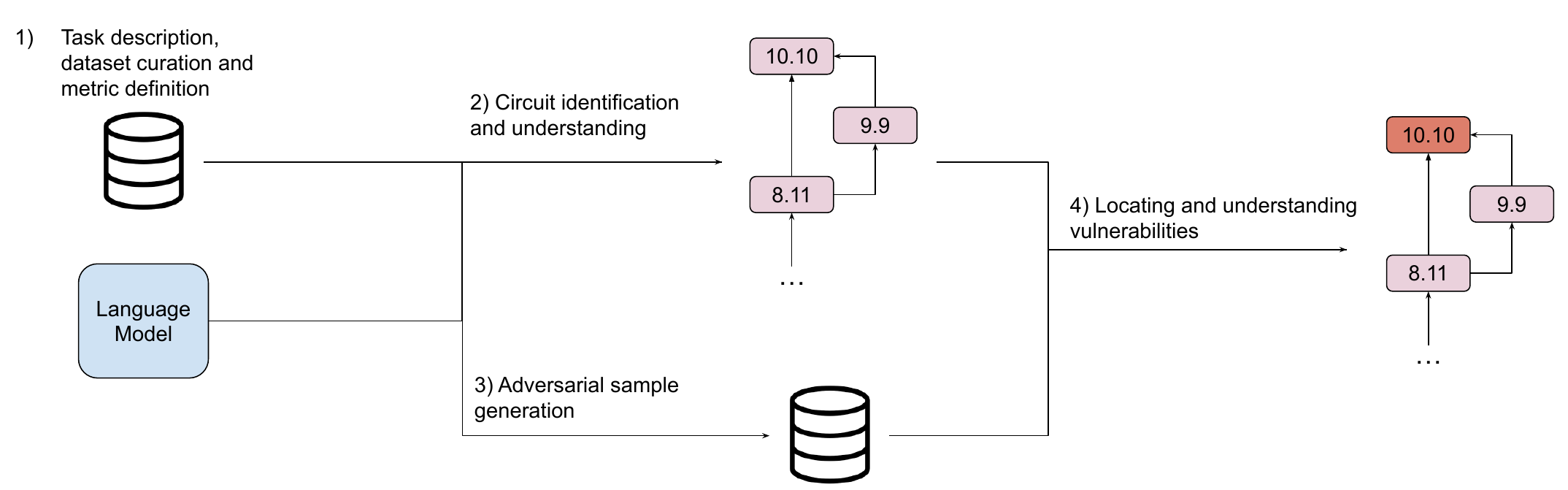}
    \caption{Summary of the workflow of our proposal to detect and locate vulnerabilities in language models.}
    \label{fig:diagram}
\end{figure*}

As previously-mentioned, the aim of our proposed approach is to be able to systematically detect and understand vulnerabilities in language models. In order to do so, our approach uses both MI and gradient-based adversarial sample generation techniques to first identify the underlying circuit associated to a given task and then locate and understand the possible vulnerabilities of the components that compose such circuit. In summary, our approach is split into the following steps, which are also shown on Fig. \ref{fig:diagram}:    

\begin{enumerate}
    \item \textbf{Task description, dataset curation and metric definition:} First, we must clearly define the behavior or task that we wish to study (e.g. acronym prediction on GPT-2 Small). Then, we build a dataset that represents such behavior of study and a metric to quantify the ability of the model to perform such task. Both the metric and the dataset will be used on the next steps to identify the underlying circuit and detect its possible vulnerabilities.
    
    \item \textbf{Circuit identification and understanding:} Then, in accordance to the typical MI workflow \cite{conmy2023towards}, we apply a series of systematic activation patching experiments to identify the underlying circuit. In addition to these experiments, different MI techniques could be used to have a basic understanding of the circuit, such as inspecting the attention patterns or weight matrices \cite{olsson2022context}.

    
    \item \textbf{Adversarial sample generation:} At this point, the circuit is identified and we have a basic understanding of how it works. Hence, the next step is to automatically generate adversarial samples that will be used to detect vulnerabilities. To this aim, we design a general gradient-based method that enables us to optimize the selected parts of a correctly classified sample so that it is misclassified, i.e. it becomes an adversarial sample. The aim of this step is two-fold: first, to give us an initial understanding of the possible vulnerabilities that the circuit may have, and second, to build an adversarial corrupted dataset that will be used on the following step to locate the exact components of the circuit that are affected.  
    

    \item \textbf{Locating and understanding vulnerabilities:} Using the previously-generated adversarial samples, we perform a series of logit attribution experiments. While the previous activation experiments were used to identify the underlying circuit, these attribution experiments will enable us to locate the components of the circuit that are affected by a vulnerability. Finally, different MI techniques can be used to understand the source of the vulnerability.  

\end{enumerate}


As a summary, the first two steps are focused on discovering and understanding the underlying circuit associated to a certain task. Once that we have narrowed our scope, the next two steps are performed to detect the possible vulnerabilities of the circuit and locate which components of the circuit are affected.

\subsection{Task Description, Dataset Curation and Metric Definition}
\label{subsec:task}

First, it is crucial to clearly define the task of study and build a dataset that elicits such behavior. For example, when studying the task of 3-letter acronym prediction, one could build a synthetic dataset composed of three words together with its corresponding acronym. It is important to remark that this dataset is \emph{not} used for training, but to perform the corresponding patching experiments that will be used to both locate the circuit and its possible vulnerabilities.

Then, we should define a metric that quantifies the ability of the model to perform the task of study. It is common among the MI literature to use the \emph{logit difference} or similar. In our case, we decide to use the following metric, which can be applied to any general task: 

\begin{equation}
\label{eq:logit_diff}
    \text{logit\_diff}_i = \text{logits}_{a_i} - \underset{a_j \in \mathcal{L} \backslash \{a_i\}}{\text{max}} \text{logits}_{a_j}
\end{equation}

where $\mathcal{L}$ is the set of possible answers (e.g. the set of capital letters on the task of acronym prediction), $a_i$ is the correct answer and $\text{logits}_{a_j}$ is the logit associated to the token $a_j$. Essentially, this metric enables us to quantify the ability of the model to perform the task of study: the higher the logit difference, the better it performs.

\subsection{Circuit Identification and Understanding}

Once that the dataset is curated and the metric is clearly defined, we will apply a series of activation patching experiments to systematically identify the underlying circuit associated to the task of study. As previously-mentioned, activation patching consists on \emph{patching} (i.e. replacing) the activations of a given component with the activations obtained by running the model on a \emph{corrupted prompt}. If patching the activation of a given component causes a large drop of performance as specified by the previously-defined metric (e.g the logit difference), it implies that such component is relevant to the task of study, hence enabling us to locate the circuit. The aim of this step is to locate a subset of the model that is responsible for the task of study, hence narrowing our focus when it comes to detecting the possible vulnerabilities. 

In addition to the activation patching experiments, different MI techniques could be used to get a basic understanding of the circuit, which could be helpful for our objective. This includes looking at the attention patterns, inspecting the weight matrices, or performing logit attribution techniques, among others \cite{olsson2022context}. 

\subsection{Adversarial Sample Generation}

Once we have identified the underlying circuit and have a basic understanding of how it works, we proceed to identify its possible vulnerabilities by first generating adversarial samples. More formally, given a neural network $f_\theta : \mathcal{X} \rightarrow Y$, and a sample $x \in \mathcal{X}$ that is correctly predicted by the model $y = f_\theta (x)$, an adversarial sample $x'$ is defined as a sample that is imperceptibly close to $x$ ($\rho(x, x') < \epsilon$, where $\rho$ is a similarity metric such as the L1 norm) but it is misclassified by the model ($y \neq f_\theta(x')$).

The process of generating adversarial samples can be formulated as an optimization problem by first defining an \emph{adversarial loss} that encourages the generation of samples that are misclassified by the model. For example, it is common to use the margin loss \cite{kaplan2020scaling}:

\begin{equation}
\begin{aligned}
    \mathcal{L}_{\text{margin}} & (x, y; f_\theta) = \\
    & \text{max} \left( f_\theta(x)_y - \underset{k \neq y}{\text{max}} f_\theta (x)_k + \kappa, 0 \right) 
\end{aligned}
\end{equation}

Once that the loss is defined, the generation process can be stated as:

\begin{equation}
    \underset{x'}{\text{min}}\, \mathcal{L}_{\text{margin}} (x', y; f_\theta) \quad \text{subject to}\, \rho(x, x') < \epsilon
\end{equation}

where $\rho$ is a similarity metric such as the L1 norm, which is commonly used on images. 

Even though the previous method is widely applied on the image and speech domains, it is not directly applicable to the text domain because the data space is discrete, therefore gradient-based methods cannot be applied. In order to overcome this, we will use a technique similar to the one that \cite{wen2023hard} used for prompt tuning and discovery. Briefly, the main idea is to optimize on the continuous embedding space, instead of the discrete text space. 

More formally, let $\mathbf{A} \in \mathcal{V}^{N}$ be an initial sample from the dataset, where $\mathcal{V}$ is the set containing all the tokens from the vocabulary and $N$ is the number of tokens of the sample. As $\mathbf{A}$ lies in a discrete space, it is not possible to directly optimize $\mathbf{A}$ so that a loss $\mathcal{L}$ is minimized. However, the sample can be embedded into the continuous embedding space $\mathbf{P} = \text{Embed}(\mathbf{A}) \in \mathbb{R}^{N \times d}$, where $d$ is the dimension of the space. As it lies on the continuous space, $\mathbf{P}$ can be optimized to minimize the previously-defined loss and then performing the inverse operation to return back to the token space, $\mathbf{A} = \text{Embed}^{-1}(\mathbf{P})$. Therefore, our method will be based on this technique to automatically generate adversarial samples that will be used to detect and understand vulnerabilities in the underlying circuit.

\begin{algorithm}[!tb]
\caption{Adversarial Sample Generation}\label{alg:adv}
\KwData{Model $f_\theta$, adversarial loss $\mathcal{L}$, vocabulary embedding $\mathbf{E}$, dataset $D$, number of steps $num\_steps$, learning rate $\alpha$, binary mask $\mathbf{m}$}
\KwResult{Generated adversarial sample $\mathbf{A}$}

\Comment{Sample A from the dataset}
Initialize $\mathbf{A} \sim D$

\Comment{Obtain the embeddings of $\mathbf{A}$}
Initialize $P \leftarrow \text{Embed}(\mathbf{A})$

\For{$i\gets1$ \KwTo $num\_steps$}{
    \Comment{Project into real embeddings}
    $\mathbf{P'} \leftarrow \text{Proj}_{\mathbf{E}}(\mathbf{P})$
    
    \Comment{Compute the gradient w.r.t projected sample}
    $\mathbf{G} = \nabla_{\mathbf{P'}} \mathcal{L} \left( \mathbf{P'}, y, f_\theta \right)$

    \Comment{Update the continuous embedding}
    $\mathbf{P} \leftarrow \mathbf{P} - \alpha \mathbf{m} \mathbf{G} $
    }

    \Comment{Project into real embeddings}
    $\mathbf{P'} \leftarrow \text{Proj}_{\mathbf{E}}(\mathbf{P})$

    \Comment{Unembed}
    $\mathbf{A} \leftarrow \text{Unembed}(\mathbf{P'})$
    
    \Return{$\mathbf{A}$}
\end{algorithm}

The pseudocode of our approach for adversarial sample generation is presented in Algorithm \ref{alg:adv}. First, an initial sample $\mathbf{A}$ is extracted from the dataset and it is embedded to obtain $\mathbf{P}$. Then, $\mathbf{P}$ is iteratively updated so that $\mathcal{L}$ is minimized by (i) projecting $\mathbf{P}$ into the nearest embedding vectors (i.e. each row of the embedding matrix) (ii) computing the gradient of $\mathcal{L}$ w.r.t the projected embeddings $\mathbf{P}'$ and (iii) using the masked gradients $\textbf{m} \textbf{G}$ to update the continuous embeddings $\mathbf{P}$. Note that the gradients are multiplied by a binary mask $\mathbf{m} \in \{0, 1\}^{N}$ that is specified by the user to control which parts of the sample are changed or remain constant.  

In our approach, the previous algorithm will be used to generate adversarial samples. For example, when looking for vulnerabilities on a palindrome classifier circuit, we can try generating non-palindromes that are incorrectly classified as palindromes by defining a loss that encourages classification error and using the algorithm to optimize different parts of the starting prompts. Notice that this process will be typically iterative, where we slowly refine the initial prompts and parts to optimize, with the objective of (i) obtaining a preliminary understanding about the possible vulnerabilities that the circuit may have (i.e. a certain position of the sequence) and (ii) prepare a dataset of adversarial samples that will be used on the next step to locate the vulnerabilities.

Finally, it is important to remark that the proposed algorithm is general and can be applied to any differentiable model that uses an embedding matrix, which is common across almost all language models. Also, our method supposes a better approach versus using a brute-force approach to generate adversarial samples (for example, by naively replacing tokens of the initial sample until it is misclassified), as the complexity of this methods grows exponentially as the sample length and vocabulary grows. 

\subsection{Locating and Understanding Vulnerabilities}

On the previous step, Algorithm \ref{alg:adv} was used to iteratively obtain adversarial samples and get an preliminary understanding about the possible vulnerabilities that the circuit of study may have. Now, these generated samples will be used to locate the components affected by the possible vulnerabilities by performing a series of logit attribution experiments. Specifically, we will compute the logit attribution of the different components of the circuit to see their individual contribution to the final prediction. If a component has a large negative logit attribution, it implies that it contributes to the misclassification of the sample, eventually enabling us to further narrow down the scope of where the vulnerability is.



Finally, the last task is to understand the identified components by applying different MI techniques, such as inspecting the attention patterns with different prompts, analyzing the weight matrices, etc. 

\section{Detecting Vulnerabilities on GPT-2 Small}
\label{sec:cases}

In this section, we will showcase our proposal by applying it to the task of acronym prediction on GPT-2 Small in order to detect possible vulnerabilities that the underlying circuit may have. The experiments were performed with both the \emph{PyTorch} \cite{paszke2019pytorch} and \emph{TransformerLens} \cite{nanda2022transformerlens} libraries by using a 6GB RTX 3060 Laptop GPU. \footnote{The code and data required to reproduce the experiments and figures, as well as the supplementary materials, can be found in \url{https://github.com/jgcarrasco/detecting-vulnerabilities-mech-interp}}

\subsection{Task Description, Dataset Curation and Metric Definition}

We will study the task of acronym prediction on GPT-2 Small \cite{radford2019language}. Specifically, given three words and the first two letters of its corresponding acronym  (e.g. \texttt{"The Chief Executive Officer (CE"}), the task of the model will be to predict the \emph{third letter} of the acronym (e.g.  \texttt{"O"}). We selected this task because it is both complex enough to showcase our proposal and serves as an illustrative example without becoming overly extensive. Following the same concept, GPT-2 Small is sufficiently large to show that our approach can be used on a real environment without the analysis becoming too extensive. It is also important to note that we could also study the task of predicting the first/second letters, but we omit them to avoid redundancy, as it does not add any extra relevant information to the case study.

The first step consists on building a dataset that elicits the task or behavior of study. Hence, we built a dataset composed by three-letter acronyms. These acronyms were built by sampling from a public list of $91000$ nouns \cite{wordlist}. In order to properly isolate the behavior of choice and provide a clearer analysis, we selected the samples whose words and respective letters of the acronym were individually tokenized, e.g. \texttt{"|The| Slam| Quick| Amp|(|S|Q|"}, where \texttt{"|"} delimits the different tokens. A more detailed explanation of the dataset building process can be found in the supplementary materials. 

We also need a metric to quantitatively assess the performance of the model at the task of study. In this case, we will use the logit difference, as defined in Equation \ref{eq:logit_diff}, where the set of possible answers is composed by the capital letters, i.e. $\mathcal{L} = \{\text{``A"}, \text{``B"},... \}$.

\subsection{Circuit Identification and Understanding}

Once that the task of study has been clearly defined and we have a dataset that elicits the behavior as well as a metric to assess the performance of the model regarding that task, we need to identify the underlying circuit, i.e. the components that are relevant to the task. As stated in the previous section, we will perform a series of activation patching experiments to identify such components. Specifically, as we are analyzing the task of predicting the third letter of the acronym, we will corrupt the third word of the initial samples by resampling them with a different word from the dataset and use the activations of this run to patch the original ones. If patching the activations of a given component causes a large drop in logit difference, it implies that such component is relevant for the task of study. 

\begin{figure}[!tb]
    \centering
    \includegraphics[width=\linewidth]{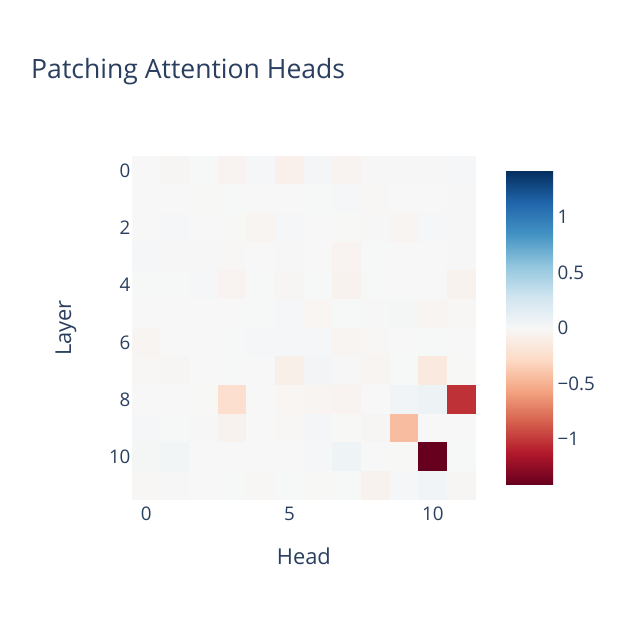}
    \caption{Variation in logit difference when patching different heads on GPT-2 Small.}
    \label{fig:patching}
\end{figure}

Figure \ref{fig:patching} shows the variation in logit difference obtained by individually patching every head of GPT-2 Small. A lower value means that patching that specific component decreases the ability of the model to perform the task, hence the component is important and forms part of the underlying circuit. The results show that heads \texttt{10.10}, \texttt{8.11} and  \texttt{9.9} cause most of the impact, so we will narrow our focus on these components and try to detect the possible vulnerabilities that may be present. 

Before getting into the next step, it is also interesting to get a basic understanding of how the selected components work. We performed a set of experiments that suggest that these heads work by attending to the word corresponding to the letter of the acronym to be predicted (i.e. the third word) and copies the capital letter of the word. These experiments can be found in the supplementary material. 

\subsection{Adversarial Sample Generation}

Once that we have identified the circuit and narrowed down our scope to a subset of components, we will generate adversarial samples with the objective of (i) discovering the possible capital letters that might be more vulnerable and (ii) using them to locate the components that are affected by the possible vulnerabilities and performing experiments to understand the source of such vulnerabilities. In order to do so, we apply Algorithm \ref{alg:adv}, setting the mask $\mathbf{m}$ so that it only optimizes the third word of the initial samples. The vocabulary embedding $\mathbf{E}$ will be composed by every possible 1-token noun that we have in our dataset. Hence, the output of this algorithm will be an adversarial sample, i.e. an acronym whose third letter is misclassified by GPT-2 Small. We repeat this procedure several times with a batch size of 128 until we obtain 1000 adversarial samples. Notice that we could gather a larger number, but the previous already had enough variety for our purposes. 

Then, we perform an analysis to study which letters are more prone to vulnerabilities. Specifically, we will compare the original probability distribution against the new generated adversarial distribution. In order to do so, if $p_{adv}^{i}$ is the probability of sampling an acronym whose third letter is the $i$th capital letter in the adversarial distribution and $p_{orig}^{i}$ is the same for the original distribution, we define the following:

\begin{equation}
 \Delta p^i = \dfrac{p_{adv}^i - p_{orig}^i}{p_{orig}^{i}}   
\end{equation}

\begin{figure}[!tb]
    \centering
    \includegraphics[width=\linewidth]{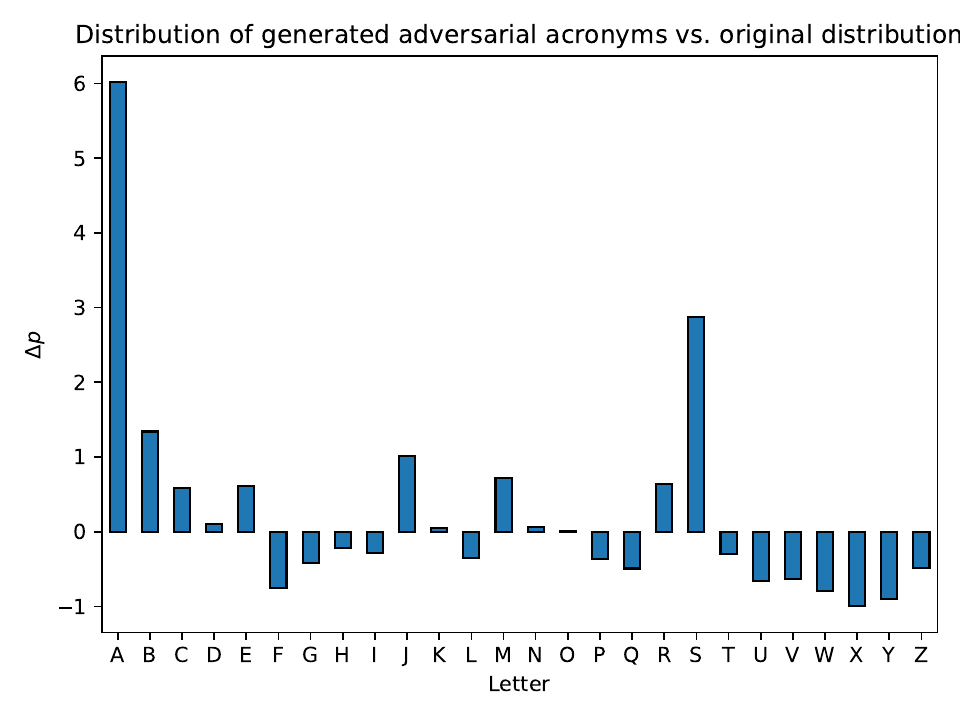}
    \caption{Distribution of the words of the dataset that begin with each letter vs. the distribution of generated adversarial acronym in terms of the initial letter of the third word.}
    \label{fig:adv_distribution}
\end{figure}

A large value of $\Delta p ^i$ implies that the $i$th capital letter is more present in the adversarial distribution as compared to the original, hence we will use it as a gauge to analyze which letters may present a vulnerability. Figure \ref{fig:adv_distribution} shows the obtained value of $\Delta p^i$ for each letter of the vocabulary. It can clearly be seen that letters \texttt{A} and \texttt{S} are 6 and 3 times more likely to appear on the generated adversarial distribution as compared to the original one, respectively. In other words, the results obtained show that these two letters are much more likely to be misclassified.

\subsection{Locating and Understanding Vulnerabilities}

As previously shown, letters \texttt{A} and \texttt{S} are considerably more prone to be misclassified, suggesting that there might be one or more components in the model that may have a vulnerability which can be exploited. 

In order to locate these components, we apply the logit attribution technique as described on Section \ref{sec:approach}. Specifically, we will cache the output of every attention head and project these vectors into the direction of the logit difference, which essentially gives us the amount that every component writes into the correct direction. Therefore, if a head outputs a negative value, it implies that it contributes to misclassifying the sample. 

\begin{figure}[!tb]
    \centering
    \includegraphics[width=\linewidth]{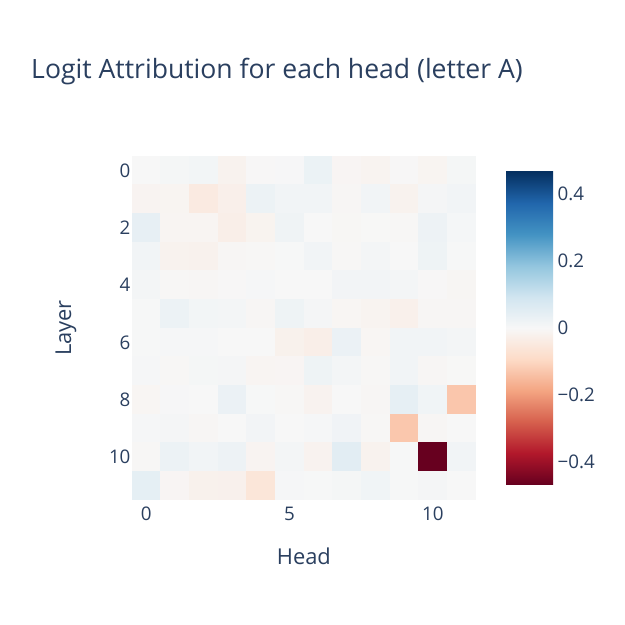}
    \caption{Logit attribution for every attention head on adversarial samples with the letter \texttt{A}. This attribution is obtained by projecting into the logit difference direction.}
    \label{fig:attribution_A}
\end{figure}

Figure \ref{fig:attribution_A} shows the logit attribution obtained on the adversarial samples with the letter \texttt{A}, which was the one that showed the largest $\Delta p$ on the previous experiment. These logit attributions reveal that the three components of the circuit that we have previously discovered contribute negatively to the output, but the contribution of head \texttt{10.10} is considerably larger than the other two. Hence, this implies that head \texttt{10.10} contributes the most when misclassifying samples with the letter \texttt{A}, which we found to be a source of vulnerability. Repeating the results for letter \texttt{S} gave us similar results which led to the same conclusion, which can be seen in the supplementary materials. 

Finally, we performed a set of logit attribution experiments to try to understand the source of such vulnerability. Specifically, we gathered the output of head \texttt{10.10} as before, but project it into the directions of the different capital letter directions. In essence, this gives us information about what this component is trying to predict. 

\begin{figure}
    \centering
    \includegraphics[width=\linewidth]{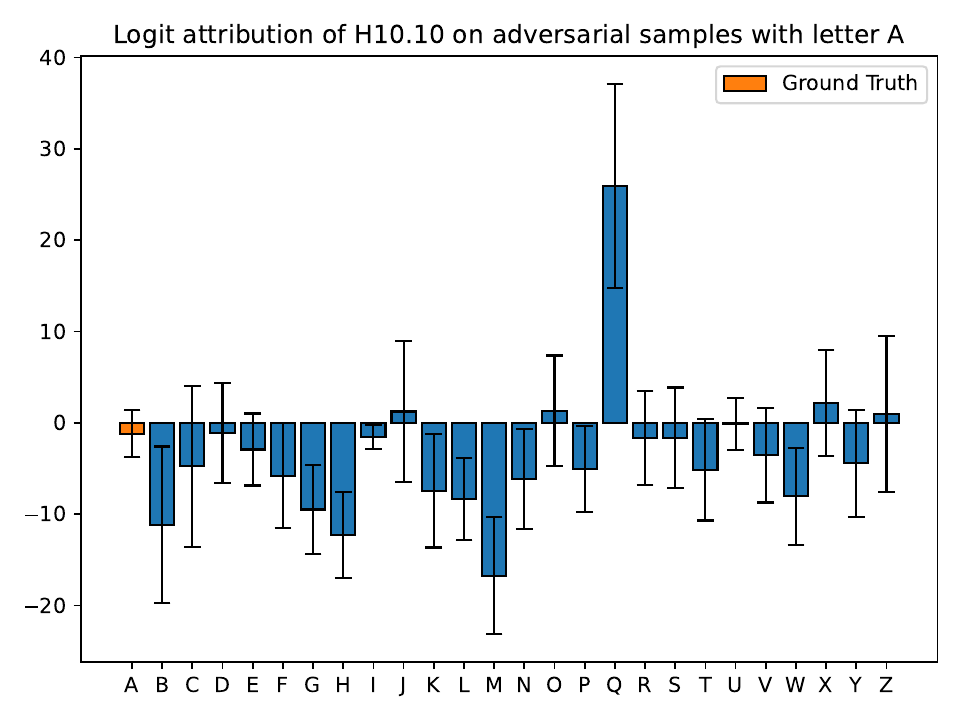}
    \caption{Logit attribution of head \texttt{10.10} on adversarial samples with the letter \texttt{A}. This attribution is obtained by projecting into the directions of the different capital letters.}
    \label{fig:A_H10_L10}
\end{figure}

Figure \ref{fig:A_H10_L10} shows the results obtained for the adversarial samples with the letter \texttt{A}. The results clearly shows that head \texttt{10.10} consistently misclassifies adversarial samples with the letter \texttt{A} by trying to predict the letter \texttt{Q}. Interestingly, repeating the experiments with the letter \texttt{S} also shows the same results. The rest of results are also included in the supplementary materials. 

\section{Discussion}
\label{sec:discussion}

The aim of the previous case study was to showcase the capabilities of our proposal when it comes to detecting vulnerabilities in language models. Specifically, we focused on looking for vulnerabilities on the task of predicting the third letter of an acronym on GPT-2 Small. The first two steps followed the typical MI workflow and enabled us to locate a few components that were responsible for the task of study, namely attention heads \texttt{10.10},  \texttt{9.9} and \texttt{8.11}. 

Then, we used the proposed Algorithm \ref{alg:adv} to automatically generate adversarial samples, i.e., samples that were misclassified by the model. This helped us to detect that the model had a strong tendency to misclassify samples whose third letter is either \texttt{A} or \texttt{S}. To locate the source of the detected vulnerabilities, we applied a series of logit attribution experiments using the previously-generated samples to check which components were contributing to such misclassification, revealing that head \texttt{10.10} was the most important source. A further analysis showed that the component had a tendency to overpredict the letter \texttt{Q} on samples whose third letter is either \texttt{A} or \texttt{S}. 

In summary, the results show that the approach can indeed be used to locate the possible vulnerabilities associated to a task and model of study, as well as analyzing what is the mechanism behind the vulnerability. Moreover, the approach is general and can be applied to any differentiable language model that uses word embeddings. 

It is also important to remark that this first version of Algorithm \ref{alg:adv} is currently focused on optimizing single-token words. Namely, optimizing two tokens belonging to a same word could lead to generating a meaningless word, as both tokens are independently modified. Nevertheless, we have decided to focus on simplicity in this initial version. As MI is a young field, this approach is already highly valuable, particularly for understanding and addressing simpler tasks prevalent in the field, making our method readily applicable in its current form. 


\section{Conclusion}
\label{sec:conclusion}

In this work, we have proposed a method aimed at detecting and locating vulnerabilities in language models from a Mechanistic Interpretability (MI) perspective. Specifically, this method enables us to detect vulnerabilities related to a concrete task by (i) obtaining the subset of the model that is responsible for that task, (ii) automatically generating adversarial samples for that task and (iii) using MI techniques together with the previous generated samples to locate the components affected by the possible vulnerability as well as understanding what is happening under the hood. In order to showcase the proposal, we applied it to locating vulnerabilities on GPT-2 Small for the task of acronym prediction, and showed that it can effectively be used to detect the possible vulnerabilities that the system may have on the task of study.


To the best of our knowledge, this is the first work that tries to locate and understand vulnerabilities from a mechanistic interpretability standpoint by zooming into the internals of the model. We believe that this perspective could help us understand why do adversarial samples exist, and eventually enable us to palliate the effects of the detected vulnerabilities.  

Regarding future work, we will focus on extending the approach so that it enables the generation of meaningful multi-token word adversarial samples by adding losses that enforce fluency or meaningful semantics, such as in \cite{guo-etal-2021-gradient}. Another research direction will be to work on methods that palliate the effects of the vulnerabilities detected with the method presented here. 

\section*{Acknowledgments}

This work has been co-funded by the BALLADEER (PROMETEO/2021/088) project, a Big Data analytical platform for the diagnosis and treatment of Attention Deficit Hyperactivity Disorder (ADHD) featuring extended reality, funded by the \emph{Conselleria de Innovación, Universidades, Ciencia y Sociedad Digital (Generalitat Valenciana)} and the AETHER-UA project (PID2020-112540RB-C43), a smart data holistic approach for context-aware data analytics: smarter machine learning for business modelling and analytics, funded by the \emph{Spanish Ministry of Science and Innovation}. Jorge García-Carrasco holds a predoctoral contract (CIACIF/2021/454) granted by the \emph{Conselleria de Innovación, Universidades, Ciencia y Sociedad Digital (Generalitat Valenciana)}.

\bibliographystyle{named}
\bibliography{ijcai24}

\begin{thebibliography}{}

\bibitem[\protect\citeauthoryear{Ba \bgroup \em et al.\egroup }{2016}]{ba2016layer}
Jimmy~Lei Ba, Jamie~Ryan Kiros, and Geoffrey~E Hinton.
\newblock Layer normalization.
\newblock {\em arXiv preprint arXiv:1607.06450}, 2016.

\bibitem[\protect\citeauthoryear{Brown \bgroup \em et al.\egroup }{2020}]{brown2020language}
Tom Brown, Benjamin Mann, Nick Ryder, Melanie Subbiah, Jared~D Kaplan, Prafulla Dhariwal, Arvind Neelakantan, Pranav Shyam, Girish Sastry, Amanda Askell, et~al.
\newblock Language models are few-shot learners.
\newblock {\em Advances in {N}eural {I}nformation {P}rocessing {S}ystems, {NeurIPS 2020}}, 33:1877--1901, 2020.

\bibitem[\protect\citeauthoryear{Cascella \bgroup \em et al.\egroup }{2023}]{cascella2023evaluating}
Marco Cascella, Jonathan Montomoli, Valentina Bellini, and Elena Bignami.
\newblock Evaluating the feasibility of chatgpt in healthcare: an analysis of multiple clinical and research scenarios.
\newblock {\em Journal of Medical Systems}, 47(1):33, 2023.

\bibitem[\protect\citeauthoryear{Chan \bgroup \em et al.\egroup }{2022}]{causalscrubbing}
Lawrence Chan, Adrià Garriga-Alonso, Nicholas Goldwosky-Dill, Ryan Greenblatt, Jenny Nitishinskaya, Ansh Radhakrishnan, Buck Shlegeris, and Nate Thomas.
\newblock Causal scrubbing, a method for rigorously testing interpretability hypotheses.
\newblock \url{https://www.alignmentforum.org/posts/JvZhhzycHu2Yd57RN/causal-scrubbing-a-method-for-rigorously-testing}, 2022.
\newblock Accessed: 2024-01-14.

\bibitem[\protect\citeauthoryear{Conmy \bgroup \em et al.\egroup }{2023}]{conmy2023towards}
Arthur Conmy, Augustine~N. Mavor-Parker, Aengus Lynch, Stefan Heimersheim, and Adri{\`a} Garriga-Alonso.
\newblock Towards automated circuit discovery for mechanistic interpretability.
\newblock In {\em Thirty-seventh Conference on Neural Information Processing Systems, {NeurIPS 2023}}, 2023.

\bibitem[\protect\citeauthoryear{Elhage \bgroup \em et al.\egroup }{2021}]{elhage2021mathematical}
Nelson Elhage, Neel Nanda, Catherine Olsson, Tom Henighan, Nicholas Joseph, Ben Mann, Amanda Askell, Yuntao Bai, Anna Chen, Tom Conerly, et~al.
\newblock A mathematical framework for transformer circuits.
\newblock {\em Transformer Circuits Thread}, 1, 2021.
\newblock https://transformer-circuits.pub/2021/framework/index.html Accessed: 2024-01-14.

\bibitem[\protect\citeauthoryear{Elhage \bgroup \em et al.\egroup }{2022}]{elhage2022toy}
Nelson Elhage, Tristan Hume, Catherine Olsson, Nicholas Schiefer, Tom Henighan, Shauna Kravec, Zac Hatfield-Dodds, Robert Lasenby, Dawn Drain, Carol Chen, et~al.
\newblock Toy models of superposition.
\newblock {\em arXiv preprint arXiv:2209.10652}, 2022.

\bibitem[\protect\citeauthoryear{Garc\'{i}a-Carrasco \bgroup \em et al.\egroup }{2024}]{pmlr-v238-garcia-carrasco24a}
Jorge Garc\'{i}a-Carrasco, Alejandro Mat\'{e}, and Juan Trujillo.
\newblock How does {GPT-2} predict acronyms? {E}xtracting and understanding a circuit via mechanistic interpretability.
\newblock In Sanjoy Dasgupta, Stephan Mandt, and Yingzhen Li, editors, {\em Proceedings of The 27th International Conference on Artificial Intelligence and Statistics}, volume 238 of {\em Proceedings of Machine Learning Research}, pages 3322--3330. PMLR, 02--04 May 2024.

\bibitem[\protect\citeauthoryear{Goodfellow \bgroup \em et al.\egroup }{2015}]{DBLP:journals/corr/GoodfellowSS14}
Ian~J. Goodfellow, Jonathon Shlens, and Christian Szegedy.
\newblock Explaining and harnessing adversarial examples.
\newblock In Yoshua Bengio and Yann LeCun, editors, {\em 3rd International Conference on Learning Representations, {ICLR} 2015, San Diego, CA, USA, May 7-9, 2015, Conference Track Proceedings}, 2015.

\bibitem[\protect\citeauthoryear{Guo \bgroup \em et al.\egroup }{2021}]{guo-etal-2021-gradient}
Chuan Guo, Alexandre Sablayrolles, Herv{\'e} J{\'e}gou, and Douwe Kiela.
\newblock Gradient-based adversarial attacks against text transformers.
\newblock In Marie-Francine Moens, Xuanjing Huang, Lucia Specia, and Scott Wen-tau Yih, editors, {\em Proceedings of the 2021 Conference on Empirical Methods in Natural Language Processing}, pages 5747--5757, Online and Punta Cana, Dominican Republic, November 2021. Association for Computational Linguistics.

\bibitem[\protect\citeauthoryear{Hanna \bgroup \em et al.\egroup }{2023}]{hanna2023does}
Michael Hanna, Ollie Liu, and Alexandre Variengien.
\newblock How does {GPT}-2 compute greater-than?: Interpreting mathematical abilities in a pre-trained language model.
\newblock In {\em Thirty-seventh Conference on Neural Information Processing Systems, {NeurIPS 2023}}, 2023.

\bibitem[\protect\citeauthoryear{Heimersheim and Janiak}{2023}]{docstring}
Stefan Heimersheim and Jett Janiak.
\newblock A circuit for {P}ython docstrings in a 4-layer attention-only transformer.
\newblock \url{https://www.alignmentforum.org/posts/u6KXXmKFbXfWzoAXn/a-circuit-for-python-docstrings-in-a-4-layer-attention-only}, 2023.
\newblock Accessed: 2024-01-14.

\bibitem[\protect\citeauthoryear{Huang \bgroup \em et al.\egroup }{2020}]{huang2020survey}
Xiaowei Huang, Daniel Kroening, Wenjie Ruan, James Sharp, Youcheng Sun, Emese Thamo, Min Wu, and Xinping Yi.
\newblock A survey of safety and trustworthiness of deep neural networks: Verification, testing, adversarial attack and defence, and interpretability.
\newblock {\em Computer Science Review}, 37:100270, 2020.

\bibitem[\protect\citeauthoryear{Kaplan \bgroup \em et al.\egroup }{2020}]{kaplan2020scaling}
Jared Kaplan, Sam McCandlish, Tom Henighan, Tom~B Brown, Benjamin Chess, Rewon Child, Scott Gray, Alec Radford, Jeffrey Wu, and Dario Amodei.
\newblock Scaling laws for neural language models.
\newblock {\em arXiv preprint arXiv:2001.08361}, 2020.

\bibitem[\protect\citeauthoryear{Liu \bgroup \em et al.\egroup }{2020}]{liu2020adversarial}
Xiaodong Liu, Hao Cheng, Pengcheng He, Weizhu Chen, Yu~Wang, Hoifung Poon, and Jianfeng Gao.
\newblock Adversarial training for large neural language models.
\newblock {\em arXiv preprint arXiv:2004.08994}, 2020.

\bibitem[\protect\citeauthoryear{Madry \bgroup \em et al.\egroup }{2018}]{madry2018towards}
Aleksander Madry, Aleksandar Makelov, Ludwig Schmidt, Dimitris Tsipras, and Adrian Vladu.
\newblock Towards deep learning models resistant to adversarial attacks.
\newblock In {\em International Conference on Learning Representations}, 2018.

\bibitem[\protect\citeauthoryear{McGrath \bgroup \em et al.\egroup }{2023}]{mcgrath2023hydra}
Thomas McGrath, Matthew Rahtz, Janos Kramar, Vladimir Mikulik, and Shane Legg.
\newblock The hydra effect: Emergent self-repair in language model computations.
\newblock {\em arXiv preprint arXiv:2307.15771}, 2023.

\bibitem[\protect\citeauthoryear{Meng \bgroup \em et al.\egroup }{2022}]{meng2022locating}
Kevin Meng, David Bau, Alex Andonian, and Yonatan Belinkov.
\newblock Locating and editing factual associations in gpt.
\newblock {\em Advances in {N}eural {I}nformation {P}rocessing {S}ystems, {NeurIPS} 2022}, 35:17359--17372, 2022.

\bibitem[\protect\citeauthoryear{Nanda and Bloom}{2022}]{nanda2022transformerlens}
Neel Nanda and Joseph Bloom.
\newblock Transformerlens.
\newblock \url{https://github.com/neelnanda-io/TransformerLens}, 2022.
\newblock Accessed: 2024-01-14.

\bibitem[\protect\citeauthoryear{Olah \bgroup \em et al.\egroup }{2020}]{olah2020zoom}
Chris Olah, Nick Cammarata, Ludwig Schubert, Gabriel Goh, Michael Petrov, and Shan Carter.
\newblock Zoom in: An introduction to circuits.
\newblock {\em Distill}, 5(3):e00024--001, 2020.

\bibitem[\protect\citeauthoryear{Olsson \bgroup \em et al.\egroup }{2022}]{olsson2022context}
Catherine Olsson, Nelson Elhage, Neel Nanda, Nicholas Joseph, Nova DasSarma, Tom Henighan, Ben Mann, Amanda Askell, Yuntao Bai, Anna Chen, Tom Conerly, Dawn Drain, Deep Ganguli, Zac Hatfield-Dodds, Danny Hernandez, Scott Johnston, Andy Jones, Jackson Kernion, Liane Lovitt, Kamal Ndousse, Dario Amodei, Tom Brown, Jack Clark, Jared Kaplan, Sam McCandlish, and Chris Olah.
\newblock In-context learning and induction heads.
\newblock {\em Transformer Circuits Thread}, 2022.
\newblock https://transformer-circuits.pub/2022/in-context-learning-and-induction-heads/index.html Accessed: 2024-01-14.

\bibitem[\protect\citeauthoryear{Paszke \bgroup \em et al.\egroup }{2019}]{paszke2019pytorch}
Adam Paszke, Sam Gross, Francisco Massa, Adam Lerer, James Bradbury, Gregory Chanan, Trevor Killeen, Zeming Lin, Natalia Gimelshein, Luca Antiga, et~al.
\newblock Pytorch: An imperative style, high-performance deep learning library.
\newblock {\em Advances in {N}eural {I}nformation {P}rocessing {S}ystems, {NeurIPS 2019}}, 32, 2019.

\bibitem[\protect\citeauthoryear{Piscitelli}{2016}]{wordlist}
Jordan Piscitelli.
\newblock Simple wordlists.
\newblock \url{https://github.com/taikuukaits/SimpleWordlists}, 2016.
\newblock Accessed: 2023-11-18.

\bibitem[\protect\citeauthoryear{Radford \bgroup \em et al.\egroup }{2019}]{radford2019language}
Alec Radford, Jeffrey Wu, Rewon Child, David Luan, Dario Amodei, Ilya Sutskever, et~al.
\newblock Language models are unsupervised multitask learners.
\newblock {\em OpenAI blog}, 1(8):9, 2019.

\bibitem[\protect\citeauthoryear{Shayegani \bgroup \em et al.\egroup }{2023}]{shayegani2023survey}
Erfan Shayegani, Md~Abdullah~Al Mamun, Yu~Fu, Pedram Zaree, Yue Dong, and Nael Abu-Ghazaleh.
\newblock Survey of vulnerabilities in large language models revealed by adversarial attacks.
\newblock {\em arXiv preprint arXiv:2310.10844}, 2023.

\bibitem[\protect\citeauthoryear{Vaswani \bgroup \em et al.\egroup }{2017}]{vaswani2017attention}
Ashish Vaswani, Noam Shazeer, Niki Parmar, Jakob Uszkoreit, Llion Jones, Aidan~N Gomez, {\L}ukasz Kaiser, and Illia Polosukhin.
\newblock Attention is all you need.
\newblock {\em Advances in {N}eural {I}nformation {P}rocessing {S}ystems, {NeurIPS 2017}}, 30, 2017.

\bibitem[\protect\citeauthoryear{Wang \bgroup \em et al.\egroup }{2019}]{wang2019improving}
Dilin Wang, Chengyue Gong, and Qiang Liu.
\newblock Improving neural language modeling via adversarial training.
\newblock In {\em International Conference on Machine Learning, {ICML 2019}}, pages 6555--6565. PMLR, 2019.

\bibitem[\protect\citeauthoryear{Wang \bgroup \em et al.\egroup }{2023}]{wang2022interpretability}
Kevin~Ro Wang, Alexandre Variengien, Arthur Conmy, Buck Shlegeris, and Jacob Steinhardt.
\newblock Interpretability in the wild: a circuit for indirect object identification in {GPT}-2 small.
\newblock In {\em The Eleventh International Conference on Learning Representations, {ICLR 2023}}, 2023.

\bibitem[\protect\citeauthoryear{Wen \bgroup \em et al.\egroup }{2023}]{wen2023hard}
Yuxin Wen, Neel Jain, John Kirchenbauer, Micah Goldblum, Jonas Geiping, and Tom Goldstein.
\newblock Hard prompts made easy: Gradient-based discrete optimization for prompt tuning and discovery.
\newblock In {\em Thirty-seventh Conference on Neural Information Processing Systems, {NeurIPS 2023}}, 2023.

\end{thebibliography}

\end{document}